\documentclass{bvm}%

\begin{document}

\selectlanguage{english}

\title{Do 3D CT Vision-language Models See the Nodule?}
\subtitle{Paired Finding Insertion as a Label-free Audit}
\titlerunning{Do 3D CT VLMs See the Nodule?}

\author{
	\fname{Niklas} \lname{Koser} \inst{1}\equalContribution \affiliation{Kiel University} \authorsEmail{niklas.koser@informatik.uni-kiel.de} \street{Christian-Albrechts-Platz} \housenumber{4} \zipcode{24118} \city{Kiel} \country{Germany} \isResponsibleAuthor,
	\fname{Tristan} \lname{Kirscher} \inst{2,3}\equalContribution \affiliation{University of Strasbourg} \authorsEmail{tristan.kirscher@unistra.fr} \street{Boulevard Sébastien Brant} \housenumber{300} \zipcode{67412} \city{Illkirch} \country{France},
	\fname{Marten} \lname{Finck} \inst{1} \affiliation{Kiel University} \authorsEmail{marten.finck@informatik.uni-kiel.de} \street{Christian-Albrechts-Platz} \housenumber{4} \zipcode{24118} \city{Kiel} \country{Germany},
	\fname{S{\"o}ren} \lname{Pirk} \inst{1} \affiliation{Kiel University} \authorsEmail{sp@informatik.uni-kiel.de} \street{Christian-Albrechts-Platz} \housenumber{4} \zipcode{24118} \city{Kiel} \country{Germany}
}
\authorrunning{Koser et al.}
\institute{
\inst{1} Visual Computing and Artificial Intelligence, Kiel University, Kiel\\
\inst{2} ICube Laboratory, CNRS UMR 7357, University of Strasbourg, Strasbourg\\
\inst{3} CLCC Institut-Strauss, Strasbourg
}
\email{niklas.koser@informatik.uni-kiel.de}

\maketitle

\begin{abstract}
	Benchmarks for 3D CT Vision-language Models (VLMs) assess agreement with radiology reports, but not whether the answer responds to the finding itself, which requires controlled image manipulation.
	We therefore insert lung nodules (our main focus), plus pleural effusions and pneumothoraces of controlled size and attenuation into public chest CTs.
	For seven public VLMs, we aks whether the finding is present in each modified CT, its original, and a sham edit of the same region, and use the log-odds of yes versus no as presence score.
	MedGemma~1.5 and the general-purpose Qwen3-VL assigned a higher presence score to more than 90\,\% of CTs with an inserted 20\,mm nodule than to the same CT without it, but separated CTs with and without such a nodule with AUCs of only 0.72 and 0.76. 
	This score increase was small relative to the score variation between patients, and a Gaussian approximation reproduced AUCs with a mean error of 0.02.
	Four models, including CT-CHAT, responded no more to inserted nodules than to shams.
    Nodules remained in their inputs, and transplanted real nodules evoked no response either.
	However, CT-CHAT achieved an AUC of 0.72 on real nodules from its training domain, which persisted after their removal.
	Real AUCs on annotated LIDC-IDRI nodules and CT-RATE effusions were rank-correlated with synthetic AUCs across 14 model--finding pairs (pooled Spearman \(\rho=0.79\)). 
    Report-based benchmarks can thus credit detection that a paired audit refutes.
\end{abstract}

\section{Introduction}
Vision-language Models (VLMs) now process whole CT volumes \cite{0000-bai2024m3d,0000-xin2026med3dvlm,0000-sellergren2026medgemma15,0000-myronenko2026nvreasonct,0000-hamamci2026ctrate} and are usually evaluated with questions derived from radiology reports \cite{0000-hamamci2026ctrate}.
Such evaluations measure agreement with the report. Yet VLMs have shown near-chance spatial reasoning \cite{0000-monon2026lostinvolume,0000-wolf2025otherleft} and correct answers without the image \cite{0000-yan2025worsethanrandom}. Agreement therefore does not show whether an answer is based on the finding in the image. In medical physics, model observers measure the detectability of known objects in known backgrounds as a function of size and contrast \cite{0000-barrett2004foundations}.
For frozen CT encoders, detection performance has been related to lesion size and contrast \cite{0000-chevli2026bigbright}.
iTRIALSPACE inserted synthetic nodules into CT volumes, evaluated VLMs on selected slices, compared host and donor contexts, and found that the host anatomy can matter more than the nodule \cite{0000-tushar2026itrialspace}.
In 2D mammography, removing annotated lesions, using a lesion-free edit as control, showed that accuracy overstates the use of lesion evidence \cite{0000-qu2026mammo}.
For VLMs that answer questions about the whole volume, a corresponding analysis has, to our knowledge, not been reported.

We propose a paired comparison as an audit, in which we insert a finding into a CT or for nodules also remove a real one and compare the model's answer with its answer to the same question for the same CT without this change. This holds the patient CT, the background, constant within each comparison, although the background can still modulate the size of a response.
Sham edits, which repeat the manipulation without the finding, together with controls of readout and slice order, separate responses to the finding from artifacts, and a Gaussian model relates the paired score changes to the patient-level AUC.
The audit requires no labels; we apply it to seven public VLMs and three findings and use annotated real data only for validation.

\section{Materials and methods}
Our study comprises three experiments. First, we insert synthetic nodules, pleural effusions and pneumothoraces into 55 CTs of the TotalSegmentator dataset (Sec.~\ref{0000-sec-synth}). Second, we transplant 30 real nodules from LIDC-IDRI into the same CTs (Sec.~\ref{0000-sec-ext}). Third, we remove the nodules from real LIDC-IDRI and CT-RATE scans and compare the AUC before and after removal (Sec.~\ref{0000-sec-ext}). The first experiment is the label-free audit. The other two experiments and unmodified scans of both datasets serve as its validation.

\subsection{Backgrounds, synthetic findings and shams}
\label{0000-sec-synth}
\begin{figure}[b]
	\centering
	\includegraphics[width=\textwidth,trim=7.2 7.2 7.2
	6.7,clip]{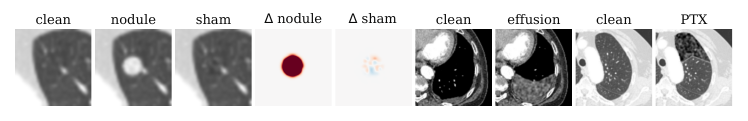}
	\caption{A test background without a finding, with a 12\,mm nodule and its sham, their differences from the clean volume (\(\pm300\)\,HU), and with a 400\,ml effusion and a 20\,\% pneumothorax.}
	\label{0000-fig-examples}
	\altText{Seven panels: three crops of the same lung region, without finding, with a bright round nodule and with the sham edit that looks like the clean crop; two difference images in which only the nodule produces a strong signal; an axial slice with a fluid crescent in the posterior left hemithorax; and an axial slice with an air collection anterior to the left lung bounded by a thin pleural line.}
\end{figure}

As backgrounds, we use CTs of the public TotalSegmentator dataset v2.0.1 \cite{0000-wasserthal2023} resampled to isotropic 1.5\,mm voxels.
Of its 1{,}228 scans, 309 cover both lungs completely and, according to the dataset metadata, show no thoracic or unclear pathology.
Each scan is cropped to the bounding box of the lungs with 15\,mm margin. 
From these scans, we randomly select 55 as backgrounds.
Nodules are inserted as spheres with a soft edge of 0.5\,mm (1.5\,mm in a second set of 12 and 20\,mm solid nodules), set to a target attenuation, blurred with a Gaussian point spread function (\(\sigma=0.6\)\,mm) and overlaid with noise that matches the surrounding lung.
We use six diameters (3, 5, 8, 12, 20, and 30\,mm), from well below the 6\,mm threshold for routine follow-up, to 30\,mm, the upper limit of a nodule \cite{0000-macmahon2017fleischner}, and four attenuations from faint ground glass to solid tissue (\(-700\), \(-500\), \(-300\), and \(+20\)\,HU).
Each center lies at least \(D/2+2\)\,mm inside the left or right lung, so that every background yields 48 lesion volumes and one clean volume.
An effusion fills the lowest part of one hemithorax with fluid of 10\,HU (50, 150,
400, and 1000\,ml) and receives noise matched to the soft tissue of the same axial slab.
A pneumothorax replaces the highest part of one lung with air behind a thin pleural line (5, 10, 20, and 35\,\% of the lung volume) and receives noise matched to the air outside the body.
For the 12\,mm and 20\,mm nodules, and for every effusion and pneumothorax, a sham edit applies the same noise with the same random seed to the same region but leaves the attenuation unchanged (Fig.~\ref{0000-fig-examples}).

\subsection{Models and answer scoring}
\label{0000-sec-models}
We evaluate seven open VLMs using their released inference code; five encode the volume in 3D.
M3D-LaMed, available with a Phi-3 or a Llama-2 language model (\(32\times256^2\) voxels) \cite{0000-bai2024m3d}, and Med3DVLM (Qwen2.5-7B, \(128\times256^2\)) \cite{0000-xin2026med3dvlm} receive the volume in a lung window (W1500/L\(-\)600).
CT-CHAT (Llama-3.1-8B, \(240\times480^2\) voxels of \(1.5\times0.75^2\)\,mm) \cite{0000-hamamci2026ctrate} and NV-Reason-CT~(Qwen3.5-4B, \(192^3\) voxels of 2\,mm) \cite{0000-myronenko2026nvreasonct} were both trained on CT-RATE.
The remaining two models encode 85 axial slices as separate images, namely MedGemma~1.5 (4B, three windows) \cite{0000-sellergren2026medgemma15} and, as a general-purpose reference model, Qwen3-VL-8B with the same instruction and a lung window \cite{0000-bai2025qwen3vl}.

For every finding we ask whether it is present (e.g.\ \elqq Is there a lung nodule in this chest CT? Answer with yes or no.\erqq) and on which side it is located, using the chat template or the training format of each model.
Instead of parsing generated text, we read the logits of the first answer token and define the presence score \(s\) as the log odds of \elqq yes\erqq{} against \elqq no\erqq{} in both capitalizations.
Because some models begin their answer with a phrase, the side score compares \elqq left\erqq{} with \elqq right\erqq{} after the fixed prefix \elqq The nodule is in the\erqq, and the presence score for M3D-LaMed with Llama-2, which begins every answer with \elqq The\erqq, is read after \elqq The answer is\erqq.
NV-Reason-CT answers without reasoning, and quantized bfloat16 logits are read in float32.

\subsection{Paired response and its relation to the AUC}
The paired change \(\Delta s=s(\text{lesion})-s(\text{clean})\) measures the response to an insertion within the same background.
The paired response \(P\) is the share of positive changes, with ties counted as one half, averaged over the two sides of each background and then over backgrounds.
We express the magnitude of the change as \(\mu/\sigma\), where \(\mu\) and \(\tau\) denote the mean and standard deviation of \(\Delta s\) and \(\sigma\) that of the clean scores across backgrounds.
By contrast, the AUC measures the separation between lesion and clean volumes at the population level.

If \(s\) and \(\Delta s\) are jointly Gaussian with covariance \(c\), then \(P\approx\Phi(\mu/\tau)\) and
\begin{equation}
	\mathrm{AUC}\approx\Phi\Bigl(\mu\big/\sqrt{2\sigma^2+\tau^2+2c}\Bigr)\approx\Phi\bigl(\mu/(\sigma\sqrt2)\bigr)
	\label{0000-eq-auc}
\end{equation}
where the second form holds for a constant shift.
\(P\) thus relates the shift to its own variability and the AUC to the variation between patients.
A small but uniform shift can therefore yield a high \(P\) together with a low AUC.
As all parameters come from the same data, the approximation is explanatory, not an independent prediction.

\(P\) is reported with Wilson intervals over the 55 backgrounds, which were wider than bootstrap intervals in all 175 nodule conditions and do not collapse at \(P=1\).
AUC intervals are bootstrapped over backgrounds.
We consider a response consistent if the lower Wilson bound of \(P\) exceeds 0.5, and as finding-specific only if the lesion also scores higher than its sham at the same position in such a majority of pairs.
A one-sided exact sign test agreed in 68 of the 70 comparisons.

\subsection{Controls for image use, readout and signal availability}
We also score each question without an image.
Readout checks assess whether \elqq yes\erqq{} or \elqq no\erqq{} is the most probable token, compare 100 free answers per model with the presence score and compare the side read before and after the fixed prefix.
Mirroring each volume must swap a correct side answer, and permuting all slices, or 16 or 4 blocks of slices whose internal order is kept, tests the use of slice order, and two alternative phrasings of each question test sensitivity to wording.
A white-noise observer quantifies the signal remaining after preprocessing as \(d'=\|\Delta\|_2/\sigma_n\) of the preprocessed difference \cite{0000-barrett2004foundations}, with \(\sigma_n\) the standard deviation of the high-pass filtered clean input in the lungs; it only shows that the signal is available.

\subsection{Transplanted nodules and real data}
\label{0000-sec-ext}
Attenuation, texture and edge width of the synthetic nodules are compared with real CT-RATE nodules.
We also transplant 30 real nodules from the LIDC-IDRI positive cases described below into the backgrounds.
Each donor contributes its largest nodule, provided that it is non-calcified, measures 8 to 30\,mm and has mostly aerated surroundings, and that the scan provides an aerated region for the sham.
Of the 30 nodules, 25 are solid. Each is cut out with the 50\,\% consensus mask of at least three readers.
The patch is resampled to 1.5\,mm, matched to the recipient's lung level and pasted into every background on both sides with a soft margin.
Its sham pastes nodule-free lung of the donor with the same mask.
The 30 donor nodules form a fixed test set: \(P\) pools all nodules and both sides within each background. A model responds consistently to a nodule if the lower Wilson bound of its \(P\) exceeds 0.5.

The LIDC-IDRI validation set \cite{0000-armato2011lidc} comprises 80 positive scans whose largest nodule, marked by at least three of four radiologists, measures at least 6\,mm (median of their diameters; 20 scans each for 6--10, 10--15, 15--20 and \(\geq\)20\,mm), and 80 negative scans in which no reader marked a nodule of 3\,mm or more.
The CT-RATE validation split \cite{0000-hamamci2026ctrate} provides 120 scans whose report mentions a nodule that a TotalSegmentator lung nodule mask supplied with CT-RATE confirms, 89 scans with neither, 79 patients with a pleural effusion in report labels and text, and 80 without.
CT-CHAT and NV-Reason-CT were trained on the CT-RATE training split, and the M3D models on M3D-Seg, which contains TotalSegmentator and LUNA16, a LIDC-IDRI subset containing 141 of our 160 scans and 23 of 30 donors.

We remove all nodules, using the reader masks for 66 LIDC-IDRI and the TotalSegmentator masks for 68 CT-RATE positives, excluding scans with masses above 30\,mm or no nodule-free site.
Each nodule plus a 1.5\,mm margin is replaced by nodule-free lung from the same scan, preserving the chest wall, and a shape-matched control edit is applied at a nodule-free site. The AUC compares these positives before and after removal with the unchanged
negatives.

\begin{table}[t]
	\caption{Finding-specific responses, i.e.\ responses consistently above the sham.
	Smallest synthetic solid nodule and smallest effusion with such a response (-- none up to 30\,mm).
	Number of the 30 transplanted LIDC-IDRI nodules and of the four pneumothorax sizes with such a response.
	AUC of real positive against negative scans before \(\to\) after their nodules were removed.
	\(^a\)Faint ground glass from 20\ts mm.}
	\label{0000-tab-summary}
	\begin{tabular*}{\textwidth}{l@{\extracolsep\fill}llllll}
		\hline
		& Smallest & Trans- & Smallest & PTX & \multicolumn{2}{l}{AUC before \(\to\) after removal} \\
		Model & nodule & plants & effusion & sizes & CT-RATE & LIDC-IDRI \\
		\hline
		M3D-LaMed & -- & 1/30 & 400\,ml & 1/4 & 0.57\,\(\to\)\,0.56 & 0.44\,\(\to\)\,0.43 \\
		M3D-LaMed L2 & -- & 0/30 & 400\,ml & 0/4 & 0.48\,\(\to\)\,0.49 & 0.56\,\(\to\)\,0.57 \\
		Med3DVLM & -- & 0/30 & 150\,ml & 3/4 & 0.56\,\(\to\)\,0.56 & 0.57\,\(\to\)\,0.56 \\
		CT-CHAT & -- & 0/30 & 50\,ml & 0/4 & 0.71\,\(\to\)\,0.71 & 0.65\,\(\to\)\,0.65 \\
		NV-Reason-CT & 5\,mm & 27/30 & 50\,ml & 4/4 & 0.93\,\(\to\)\,0.82 & 0.88\,\(\to\)\,0.60 \\
		MedGemma~1.5 & 12\,mm & 17/30 & 50\,ml & 0/4 & 0.76\,\(\to\)\,0.70 & 0.71\,\(\to\)\,0.60 \\
		Qwen3-VL & 12\,mm\(^a\) & 14/30 & 400\,ml & 4/4 & 0.56\,\(\to\)\,0.56 & 0.59\,\(\to\)\,0.51 \\
		\hline
	\end{tabular*}
\end{table}

\section{Results}

\subsection{Synthetic nodules}
\begin{figure}[b]
	\centering
	\includegraphics[width=.8\textwidth,trim=7.5 9.5 9.3 7.2,clip]{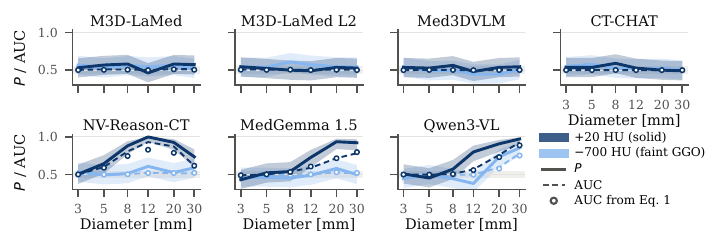}
	\caption{Nodules: paired response \(P\) (solid, 95\,\% Wilson band), observed AUC (dashed) and AUC from Eq.~\ref{0000-eq-auc} (circles) over diameter for faint ground glass and solid nodules.}
	\label{0000-fig-results}
	\altText{Seven line-plot panels versus nodule diameter. M3D-LaMed with Phi-3 and with Llama-2, Med3DVLM and CT-CHAT stay near 0.5 for both measures. NV-Reason-CT reaches a paired response of 1.0 and an AUC of 0.93 at 12 millimeters and declines at 30 millimeters. MedGemma and Qwen3-VL reach paired responses above 0.9 at 20 millimeters while their AUCs rise only to about 0.7 to 0.8; Qwen3-VL also responds to faint ground glass at 20 and 30 millimeters.}
\end{figure}

The nodule shams alone raised no model's score consistently.
Four of the seven models, both M3D-LaMed variants, Med3DVLM and CT-CHAT, showed no consistent response to inserted nodules in any condition (\(P\) 0.44 to 0.61, AUC 0.50 to 0.52, \(|\mu|\leq0.04\,\sigma\); Tab.~\ref{0000-tab-summary}), and their score changes did not exceed those after the sham~(Fig.~\ref{0000-fig-results}).
The preprocessed inputs still contained the nodules: from 8\,mm on, \(d'\) exceeded 3 for the M3D models and Med3DVLM and 40 for CT-CHAT.
Neither rephrasing nor the language model of M3D-LaMed changed this.

The other three models raised their scores.
NV-Reason-CT responded consistently from 5\,mm on (Tab.~\ref{0000-tab-summary}) and raised its score for 12\,mm solid nodules in all 55 backgrounds (\(P=1.00\) [0.93, 1.00], \(\mu=1.35\,\sigma\), AUC 0.93).
MedGemma~1.5 and Qwen3-VL raised their score for 94 and 91\,\% of the 20\,mm nodules.
Their mean increases were 0.81 and \(0.87\,\sigma\), and their AUCs were 0.72 and 0.76.
Only Qwen3-VL responded consistently to faint ground glass, a common sign of early adenocarcinoma, from 20\,mm on (\(P\) 0.74 and 0.85), although from 12\,mm on it was present in all inputs (\(d'\geq6\)).

In the 71 synthetic conditions with a consistent response, the shift form (Eq.~\ref{0000-eq-auc}) deviated from the observed AUC by 0.020 on average (maximum 0.10; circles in Fig.~\ref{0000-fig-results}), and \(\Phi(\mu/\tau)\) reproduced \(P\) within 0.03.
The largest deviations occurred for NV-Reason-CT, whose score changes grew as the clean score fell (\(r=-0.92\)), consistent with its AUC exceeding the approximation.

\subsection{Effusions and pneumothoraces}
\begin{figure}[b]
	\centering
	\includegraphics[width=.8\textwidth,trim=7.5 9.5 9.3 7.2,clip]{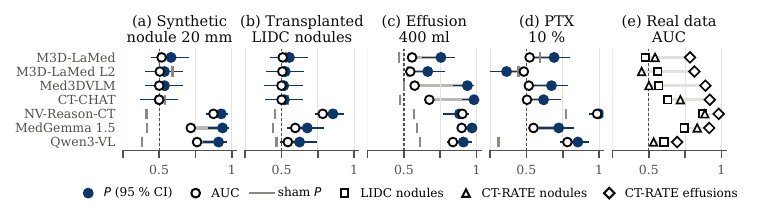}
	\caption{(a--d) \(P\) (95\,\% Wilson interval), AUC and the sham's \(P\) for 20\,mm solid and transplanted LIDC-IDRI nodules, 400\,ml effusions and 10\,\% pneumothoraces (PTX); (e) real-data AUC.}
	\label{0000-fig-dumbbell}
	\altText{Five dumbbell panels with one row per model. For synthetic 20 millimeter nodules and for transplanted LIDC-IDRI nodules, M3D-LaMed variants, Med3DVLM and CT-CHAT have paired responses and AUCs near 0.5; NV-Reason-CT has both near 0.9 and 0.8; MedGemma and Qwen3-VL have higher paired responses than AUCs, lower for the transplanted nodules. For effusions, all models have high paired responses, while the AUCs of the M3D variants, Med3DVLM and CT-CHAT stay between 0.54 and 0.67. For pneumothorax, only NV-Reason-CT and Qwen3-VL lie clearly above their sham ticks. Real data: NV-Reason-CT is highest throughout; CT-CHAT separates CT-RATE nodules better than LIDC nodules; all models separate real effusions.}
\end{figure}

Of the effusion shams, only that of Qwen3-VL at 150\,ml raised a score consistently, by \(0.06\,\sigma\).
All models scored effusions higher than their shams, from 50 to 400\,ml~(Tab.~\ref{0000-tab-summary}).
At 400\,ml, the M3D models, Med3DVLM and CT-CHAT shifted their score by only 0.14--\(0.63\,\sigma\) (AUC 0.54--0.67), MedGemma~1.5, NV-Reason-CT and Qwen3-VL by 1.4--\(1.8\,\sigma\) (AUC 0.84--0.90; Fig.~\ref{0000-fig-dumbbell}).

For pneumothoraces, the sham alone, which adds noise but no air, raised the scores of NV-Reason-CT (\(P\leq0.77\)), MedGemma~1.5 (\(P\leq0.73\)) and, at 35\,\%, Med3DVLM, and it lowered that of Qwen3-VL (\(P\) down to 0.15).
Only NV-Reason-CT and Qwen3-VL exceeded the sham at every size (Tab.~\ref{0000-tab-summary}); without it, MedGemma~1.5 would have appeared to respond to pneumothoraces.

\subsection{Image use and readout}
The scores of all models depended on the image: their clean scores varied between backgrounds, and without an image five of them produced scores outside this range~(\(|z|\) 2.2--14), so the missing response of four models does not reflect a disregard of the image.
Yes or no was the most probable token in 98--100\,\% of cases (M3D-LaMed L2 after its prefix), and free answers agreed with the score in 94--100\,\%.
The side depended on the readout: read at the first token, MedGemma~1.5 seemed to answer \elqq right\erqq{} for 92\,\% of the volumes, whereas after the prefix it localized 83\,\% of nodules \(\geq\)20\,mm and 95\,\% of effusions.
Qwen3-VL localized these nodules but mirrored the sides, naming the wrong side for 89\,\%, and swapping it for 92\,\% of mirrored volumes.
Only NV-Reason-CT depended on slice order, losing its response when single slices were permuted (\(P\) 0.53--0.58).

\subsection{Transplanted nodules and real data}
The synthetic nodules matched 33 non-calcified CT-RATE nodules in interior attenuation and texture but had a narrower edge (2.2 vs.\ 3.5\,mm); with the wider edge, \(P\) and AUC changed by at most 0.13 and 0.06.
Transplanted LIDC-IDRI nodules reproduced this division (Fig.~\ref{0000-fig-dumbbell}b).
The four non-responders showed no consistent response (\(P\) 0.52--0.55, AUC \(\leq0.51\)), whereas NV-Reason-CT, MedGemma~1.5 and Qwen3-VL responded to 27, 17 and 14 of the 30 nodules (Tab.~\ref{0000-tab-summary}).
MedGemma~1.5 and Qwen3-VL responded mainly to larger nodules; all three responded less strongly on average than to synthetic nodules of similar size, so the synthetic audit may overestimate responses to real nodules.

On LIDC-IDRI (Fig.~\ref{0000-fig-dumbbell}e), NV-Reason-CT reached an AUC of 0.87 [0.81, 0.92] and MedGemma~1.5 one of 0.74 [0.67, 0.82]; the other five models reached 0.48 to 0.63, including the M3D models, whose segmentation training data contained most of these scans.
CT-CHAT reached 0.72 on real nodules from its training domain CT-RATE and 0.63 on LIDC-IDRI, although it had not responded to any inserted nodule.
Nodule removal lowered the scores of NV-Reason-CT and MedGemma~1.5 more than the control edit in 79--86\,\% and reduced their AUCs, with larger reductions on LIDC-IDRI (Tab.~\ref{0000-tab-summary}).
For CT-CHAT, removal changed neither the scores (median \(0.00\,\sigma\)) nor the AUC, and the control edit changed no AUC by more than 0.02.

Across the 14 model--finding combinations, synthetic AUCs (mean over 8--20\,mm solid nodules and 150--1000\,ml effusions) and real AUCs were rank-correlated~(pooled Spearman \(\rho=0.79\), within findings 0.57).

\section{Discussion}
A consistent response does not imply good separation of patients. MedGemma~1.5 and Qwen3-VL raised their score for over 90\,\% of the 20\,mm nodules but reached AUCs of only 0.72 and 0.76, because the increase was small relative to the score variation between patients (Eq.~\ref{0000-eq-auc}).

Four models showed no consistent score increase for nodules present in their inputs.
CT-CHAT's AUC on real nodules from its training domain, which a report-based benchmark
would credit as detection, persisted after the nodules were removed, unlike that
of NV-Reason-CT. Whether CT-CHAT uses other image content, such as accompanying findings,
remains untested.

Evaluations should compare responses with a sham, as the pneumothorax sham alone changed four models' scores.
Scoring should use a validated readout position and side convention, as the first token was biased, one model gave no direct answer and another mirrored the sides.
Reports should give the size of a response with its consistency, as \(P\) alone can overstate patient separation.

Our work has several limitations. The synthetic findings lack associated pathology such as atelectasis, the transplanted nodules exclude juxtapleural and calcified ones, and removal relies on possibly incomplete masks.
Real negatives may contain unreported findings, Qwen3-VL's input format is ours, and all models were probed zero-shot; code, scores and \(\Delta s\) distributions will be released.

Paired comparisons with the clean CT and a sham thus show which findings change the scores of the tested volumetric VLMs, and the size of this change relative to between-patient variation largely determined patient separation.

\begin{acknowledgement}
This work was supported by the IDIR-Project (Digital Implant Research), a cooperation financed by Kiel University, University Hospital Schleswig-Holstein and Helmholtz Zentrum Hereon.
\end{acknowledgement}

\printbibliography

\end{document}